\documentclass{article}

\usepackage{PRIMEarxiv}

\usepackage[utf8]{inputenc} 
\usepackage[T1]{fontenc}    
\usepackage{hyperref}       
\usepackage{url}            
\usepackage{booktabs}       
\usepackage{amsfonts}       
\usepackage{nicefrac}       
\usepackage{microtype}      
\usepackage{fancyhdr}       
\usepackage{graphicx}       

\usepackage{amssymb}
\usepackage{amsmath}
\usepackage[flushleft]{threeparttable}
\usepackage{tabularx}
\usepackage{algorithm}
\usepackage[noend]{algpseudocode}

\title{DynaGuide: A Generalizable Dynamic Guidance Framework for Unsupervised Semantic Segmentation}

\author{
  Boujemaa Guermazi\thanks{Corresponding author.} \\
  Electrical and Computer Engineering \\
  Toronto Metropolitan University \\
  Toronto, Ontario, Canada \\
  \texttt{bguermazi@torontomu.ca} \\
  \And
  Riadh Ksantini \\
  Computer Science \\
  University of Bahrain \\
  Zallaq, Sakhir, Kingdom of Bahrain \\
  \texttt{rksantini@uob.edu.bh} \\
  \And
  Naimul Khan \\
  Electrical, Computer, and Biomedical Engineering \\
  Toronto Metropolitan University \\
  Toronto, Ontario, Canada \\
  \texttt{n77khan@torontomu.ca} \\
}

\begin{document}
\maketitle

\begin{abstract}
Unsupervised image segmentation is a critical task in computer vision. It enables dense scene understanding without human annotations, which is especially valuable in domains where labelled data is scarce. However, existing methods often struggle to reconcile global semantic structure with fine-grained boundary accuracy. This paper introduces DynaGuide, an adaptive segmentation framework that addresses these challenges through a novel dual-guidance strategy and dynamic loss optimization. 
Building on our previous work, DynaSeg, DynaGuide combines global pseudo-labels from zero-shot models such as DiffSeg or SegFormer with local boundary refinement using a lightweight CNN trained from scratch. This synergy allows the model to correct coarse or noisy global predictions and produce high-precision segmentations. At the heart of DynaGuide is a multi-component loss that dynamically balances feature similarity, Huber-smoothed spatial continuity, including diagonal relationships, and semantic alignment with the global pseudo-labels. Unlike prior approaches, DynaGuide trains entirely without ground-truth labels in the target domain and supports plug-and-play integration of diverse guidance sources.
Extensive experiments on BSD500, PASCAL VOC2012, and COCO demonstrate that DynaGuide achieves state-of-the-art performance, improving mIoU by 17.5\% on BSD500, 3.1\% on PASCAL VOC2012, and 11.66\% on COCO. With its modular design, strong generalization, and minimal computational footprint, DynaGuide offers a scalable and practical solution for unsupervised segmentation in real-world settings. Code available at \url{https://github.com/RyersonMultimediaLab/DynaGuide}
\end{abstract}

\keywords{Unsupervised image segmentation \and self-attention \and dynamic pseudo-labeling \and dual guidance \and CNN-Transformer hybrid \and adaptive loss function \and attention-guided segmentation}

\section{Introduction}
\label{Intro}



Semantic segmentation aims to assign a category label to every pixel in an image, enabling a fine-grained understanding of visual scenes. While supervised methods have achieved remarkable performance, they rely on large-scale annotated datasets, which are costly and labour-intensive to create. This limitation has driven growing interest in unsupervised segmentation, where models are trained without ground-truth labels, an essential capability in domains such as medical imaging, satellite analysis, and autonomous systems.

Despite notable progress, most unsupervised methods struggle to balance global semantic consistency with precise spatial delineation. Clustering-based techniques and contrastive learning approaches often rely on fixed loss functions or static pseudo-labels, limiting their adaptability during training. 

Recent self-supervised and zero-shot learning approaches have further advanced the field, utilizing pre-trained vision models to generate semantic features. Notably, diffusion-based models like DiffSeg have demonstrated remarkable zero-shot segmentation capabilities by extracting attention maps from stable diffusion models. However, these methods often produce coarse segmentations with unclear boundaries, limiting their effectiveness in fine-grained segmentation scenarios.
 \begin{figure*}[ht]
    \centering
    \includegraphics[width=0.95\textwidth]{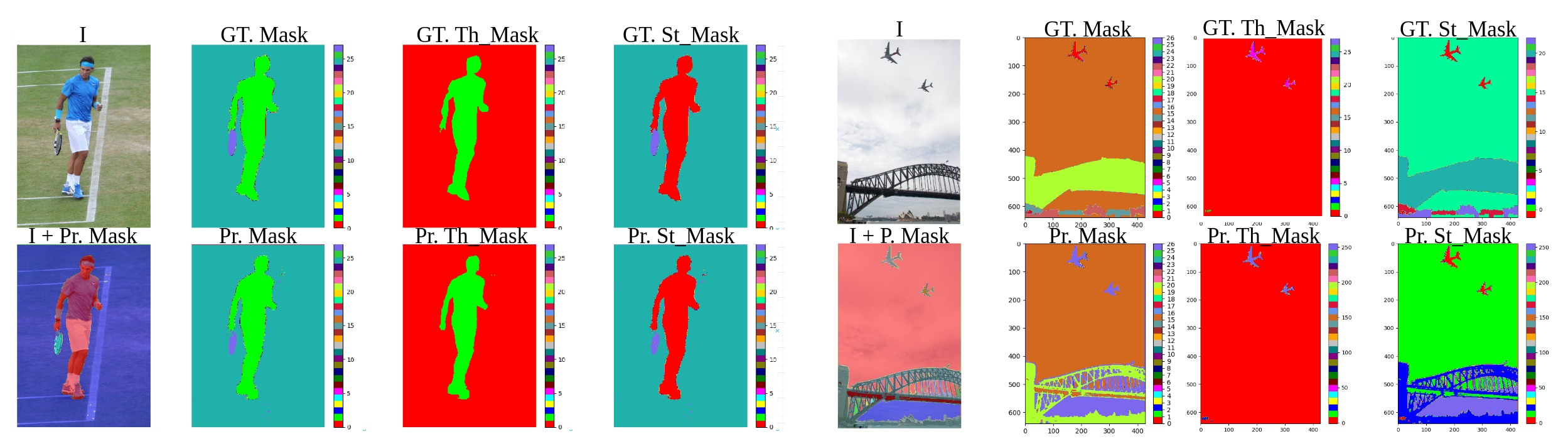} 
    \caption{Illustration of the \textit{DynaGuide} segmentation approach, showcasing the Original Image (I), Ground Truth (GT) Mask, and Predicted (Pr) outputs. The results highlight the model’s ability to segment "Things" (Th) and "Stuff" (St) categories, emphasizing its robustness in handling complex visual scenes.}
    
    \label{fig:DynaGuideimageheader}
\end{figure*}

To overcome these challenges, we propose DynaGuide, a flexible and generalizable framework that refines segmentation predictions through global-to-local dynamic guidance. DynaGuide employs a dual pseudo-label mechanism, combining global segmentation priors from external models with a lightweight convolutional neural network (CNN) that refines spatial accuracy. The model optimizes feature similarity, spatial continuity using Huber-based smoothing, and alignment with the global pseudo-labels through a novel adaptive multi-component loss function. This asymmetric dual-guidance process enhances boundary precision while maintaining semantic coherence.

A key strength of DynaGuide is its flexibility. While it operates in a fully unsupervised setting using DiffSeg for zero-shot segmentation, it also generalizes effectively to segmentation priors from supervised-pretrained models like SegFormer without requiring fine-tuning. Although SegFormer was originally trained with labeled data, it is used here solely to generate coarse semantic priors on unseen target datasets, maintaining DynaGuide’s unsupervised training protocol without accessing ground-truth annotations. By employing a lightweight CNN at inference, DynaGuide ensures computational efficiency, making it suitable for real-time and resource-constrained applications.

Our experiments show that DynaGuide achieves state-of-the-art performance, with significant improvements in mean Intersection over Union (mIoU) across benchmark datasets. The model effectively captures fine-grained details and accurately segments complex scenes. Figure~\ref{fig:DynaGuideimageheader} illustrates the DynaGuide segmentation approach, showcasing the Original Image (I) from the COCO dataset, Ground Truth (GT) Mask, and Predicted (Pr) outputs. The results highlight the model’s ability to segment “Things” (Th) and “Stuff” (St) categories, emphasizing its robustness in handling complex visual scenes.
In summary, our key contributions are:

\begin{itemize}
    \item \textbf{Generalizable Dual-Guidance Framework:} A hybrid CNN-Transformer architecture that integrates global pseudo-labels with local CNN refinement for improved segmentation accuracy.
    \item \textbf{Adaptive Multi-Component Loss Function:} A dynamic loss that balances feature similarity, spatial continuity, and global guidance alignment.
    \item \textbf{Unsupervised and Modular Guidance:} DynaGuide effectively utilizes unsupervised segmentation with DiffSeg and offers a plug-and-play capability to adapt to supervised priors like SegFormer without fine-tuning, while maintaining its unsupervised training protocol.
    \item \textbf{State-of-the-Art Performance:} Significant improvements over existing unsupervised methods across multiple datasets, with efficient inference using a lightweight CNN.  
\end{itemize}
The remainder of this paper is organized as follows: Section \ref{relatedwork} discusses related work, including recent advances in unsupervised segmentation and pseudo-labeling techniques. Section \ref{methodology} presents the detailed architecture of DynaGuide, including its dual-guidance mechanism and adaptive loss function. Section \ref{experiment} provides extensive experimental results and comparisons with state-of-the-art methods. Finally, Section \ref{conclusion} concludes the paper and highlights potential future directions.

\section{Related Work}
\label{relatedwork}
\subsection{Unsupervised Semantic Segmentation}
Unsupervised segmentation aims to group pixels into semantically coherent regions without access to annotated labels. Traditional clustering-based methods, such as K-means~\cite{macqueen1967some}, mean shift~\cite{comaniciu2002mean}, and graph-based segmentation~\cite{felzenszwalb2004efficient}, laid foundational principles for grouping pixels by appearance and proximity. However, these methods are limited by handcrafted features, sensitivity to noise, and lack of semantic abstraction.

Deep learning-based unsupervised segmentation has advanced the field by leveraging learned feature representations~\cite{IIC, caron2018deep}. Notable methods include Invariant Information Clustering (IIC)\cite{IIC} and PiCIE\cite{cho2021picie}, which use mutual information and equivariance constraints to guide pixel-level grouping. 
More recent techniques have exploited self-supervised vision models for clustering~\cite{van2021unsupervised, zadaianchuk2022unsupervised}. STEGO~\cite{hamilton2022unsupervised} and DeepSpectral~\cite{melas2022deep} extract object-level semantics from DINO-ViT features using contrastive mechanisms, while COMUS~\cite{zadaianchuk2022unsupervised} introduces attention-based refinement. While effective, these methods typically rely on fixed label assignments or post-hoc clustering, lacking mechanisms for iterative refinement or structured label evolution.

DynaGuide builds on these advances by introducing a dynamic training loop that unifies global semantic priors and local feature refinement. Unlike previous methods, it allows iterative improvement of segmentation maps through continuous feedback, producing results that are both semantically coherent and spatially precise.



\subsection{Pseudo-Labeling and Guidance Strategies}
Pseudo-labeling has become a central strategy in unsupervised and semi-supervised segmentation. Early works such DeepCluster~\cite{caron2018deep} and DIC~\cite{zhou2020dic}, followed a cluster-and-train approach in which pseudo-labels are generated through clustering and then treated as fixed supervision targets. However, this static nature often results in confirmation bias and limited adaptability to evolving representations.

More recent models have proposed dynamic pseudo-label refinement strategies. Differentiable frameworks like DFC~\cite{kim2020unsupervised} optimize clustering objectives jointly with feature learning, while methods such as IIC~\cite{IIC}, DenseSiam~\cite{zhang2022dense}, and ProtoCon~\cite{li2023acseg} improve pseudo-label quality by enforcing consistency across augmentations or across prototype assignments. Some approaches further incorporate uncertainty modeling~\cite{seong2023leveraging} or semantic alignment~\cite{ziegler2022self} to stabilize predictions over time.

While pseudo-label refinement has improved significantly in semi-supervised learning and domain adaptation, most frameworks assume access to labelled data, supervised source-domain pretraining, or ensemble-based guidance mechanisms. In related domains, methods such as DUMM~\cite{qiu2024dual} (semi-supervised segmentation), GLoDe~\cite{ding2024improving}, and Pseudo MAE~\cite{nandam2024pseudo} explore dual-guidance or multi-teacher strategies. However, these approaches are typically applied to classification or cross-domain transfer tasks and rely on either labelled data or supervised pretraining. Moreover, they generally employ multi-stage optimization rather than unified training with interdependent guidance streams.

DynaGuide sets itself apart from prior pseudo-labeling methods by employing a structured dual-guidance mechanism that ensures adaptive refinement through continuous feedback. Unlike existing works that typically operate in a single feedback loop or apply multi-stage refinement, DynaGuide integrates fixed global pseudo-labels and evolving local predictions into a unified training framework. This duality is enforced through a dynamic loss function that aligns semantic consistency with spatial precision. Crucially, the framework operates in a fully unsupervised setting when paired with DiffSeg~\cite{tian2024diffuse} as the global prior. It maintains label-free training even when integrating pre-trained models like SegFormer~\cite{xie2021segformer}— by applying them only to unseen datasets. This design enables robust segmentation without reliance on any ground-truth labels or supervised pretraining, setting DynaGuide apart from both unsupervised and semi-supervised alternatives.

\subsection{Hybrid Architectures and Diffusion-Based Guidance}
Recent advances in semantic segmentation have shown that combining local feature extraction with global context modelling can significantly improve performance. CNNs remain the backbone of many segmentation frameworks due to their efficiency and ability to preserve fine-grained spatial information. However, their limited receptive field restricts the modelling of long-range dependencies, which are essential for capturing semantic coherence in complex scenes.
Transformer-based architectures address this limitation by using self-attention mechanisms to model global interactions. Models like TransUNet~\cite{chen2021transunet}, Swin-Unet~\cite{cao2022swin}, and SegFormer~\cite{xie2021segformer} exemplify this synergy, achieving state-of-the-art results in supervised settings. However, these approaches rely heavily on labelled data for training, limiting their applicability in fully unsupervised scenarios.

Diffusion-based methods have recently emerged as powerful tools for unsupervised segmentation. DiffSeg~\cite{tian2024diffuse} leverages attention maps from stable diffusion models to produce zero-shot segmentation masks without requiring labeled data or training. These masks capture global semantic structures but often lack boundary precision and spatial consistency, necessitating additional refinement.

DynaGuide builds on this insight by integrating hybrid CNN-Transformer guidance in a fully unsupervised segmentation framework. Specifically, it uses global pseudo-labels from an external segmentation prior—such as DiffSeg or SegFormer applied to unseen datasets—to guide the learning of a lightweight CNN trained from scratch. The global priors provide semantic layout, while the CNN progressively refines spatial structure and boundary accuracy. This dual-source guidance is integrated through a dynamic training loop that does not require any ground-truth labels in the target domain.

Importantly, DynaGuide preserves its unsupervised nature when using DiffSeg as the prior and maintains modularity by generalizing across domains and segmentation priors. This allows the framework to adapt to diverse scenarios while maintaining a unified, end-to-end learning strategy. To the best of our knowledge, DynaGuide is the first to explicitly combine transformer-derived global context with CNN-based local refinement in a fully unsupervised dual-guidance framework for semantic segmentation.

In addition to transformer and diffusion-based models, the recent Segment Anything Model (SAM)~\cite{kirillov2023segment} has further pushed the boundaries of segmentation by enabling prompt-driven, zero-shot generalization across diverse domains. SAM leverages extensive supervised pretraining and powerful image-text embeddings to support segmentation from points, boxes, or masks, making it highly flexible and interactive. However, SAM requires explicit user input at inference and does not operate in a fully autonomous manner. In contrast, DynaGuide is designed for fully unsupervised settings, requiring no prompts, annotations, or supervision in the target domain. Even when using SegFormer or DiffSeg as guidance sources, DynaGuide maintains a label-free training regime and performs segmentation autonomously at inference.

\section{Methodology}
\label{methodology}

DynaGuide is a fully unsupervised segmentation framework designed to achieve robust semantic segmentation by integrating global contextual guidance with local boundary refinement. It introduces a hybrid dual-guidance approach that combines the strengths of a global segmentation model and a convolutional neural network (CNN). Unlike traditional segmentation methods that rely on labeled data, DynaGuide leverages external pseudo-labels for global structure awareness while iteratively refining boundaries using a dynamic loss function.

\subsection{Overview of the DynaGuide Framework}

\begin{figure*}[t]
    \centering
    \includegraphics[width=\textwidth]{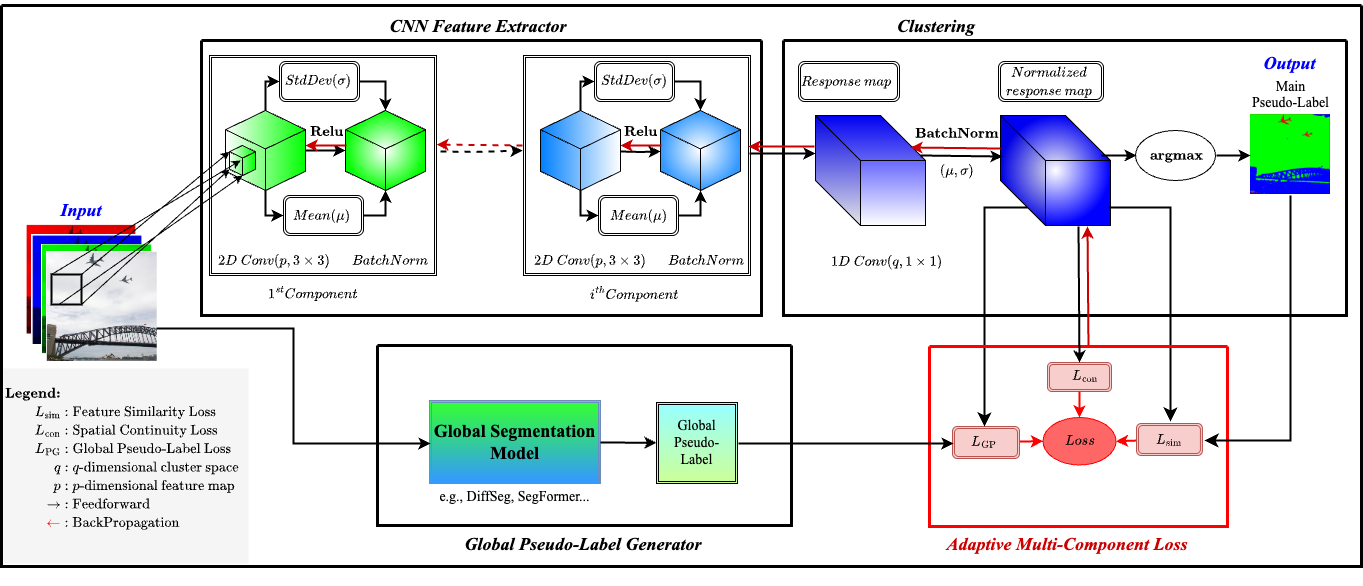}
    \caption{Architecture of DynaGuide for unsupervised image segmentation. The model uses a CNN Feature Extractor with convolutional layers, ReLU activation, and Batch Normalization (denoted by Mean (\( \mu \)) and Std Dev (\( \sigma \)) boxes) to generate \( p \)-dimensional feature map. A linear classifier and normalization produce a Normalized Response Map, which is clustered to generate the final segmentation. The Global Pseudo-Label Generator provides external pseudo-labels using a global segmentation model (e.g., DiffSeg or SegFormer) to guide the segmentation process. The Adaptive Multi-Component Loss combines Feature Similarity Loss (\( L_{\text{sim}} \)), Spatial Continuity Loss (\( L_{\text{con}} \)) using Huber Loss with diagonal components, and Global Pseudo-Label Guidance Loss (\( L_{\text{GP}} \)) for iterative refinement.}
    \label{Dynaguide_DiffSeg}
\end{figure*}

At the core of DynaGuide is the iterative enhancement of segmentation predictions through a synergy of global guidance and local feature extraction. As illustrated in figure~\ref{Dynaguide_DiffSeg}, the framework consists of three primary components:

\begin{enumerate}
\item{\textbf{Global Pseudo-Label Guidance:} 
A diffusion-based model such as DiffSeg generates static pseudo-labels through zero-shot segmentation, utilizing attention maps to capture high-level semantic structures. These global pseudo-labels offer an initial coarse segmentation that serves as a structural prior during training. Importantly, the framework supports flexibility in adopting other global segmentation models like SegFormer.}

\item{\textbf{Local Refinement:} 
A CNN is responsible for learning localized spatial features, enhancing boundary precision, and correcting segmentation inconsistencies. Through iterative learning, the CNN adapts to regions where the global guidance is inaccurate, effectively refining the segmentation mask. Residual learning mechanisms further facilitate the capture of fine details.}

\item{\textbf{Adaptive Loss Optimization:} 
The training process is governed by a dynamic, multi-component loss function that continuously adjusts the model parameters. This function integrates three key objectives: maintaining feature similarity, ensuring spatial continuity using Huber loss with diagonal regularization, and aligning local predictions with the global pseudo-labels. At inference, the model relies solely on the refined CNN predictions, achieving efficient, fully unsupervised segmentation.}
\end{enumerate}

The detailed segmentation process is outlined in Algorithm~\ref{alg:dynaguide}. During each iteration, the CNN extracts multi-scale features, generates response maps $x$, and assigns cluster labels $\mathbf{C}$. The loss function is computed to optimize the CNN parameters using gradient descent. This iterative refinement continues until convergence, ensuring robust segmentation results.

\begin{algorithm}[H]
\caption{DynaGuide Unsupervised Image Segmentation}
\label{alg:dynaguide}
\begin{algorithmic}[1]

\Require $I \in \mathbb{R}^{H \times W \times 3}$ \Comment{Input image}
\Require $T$ \Comment{Number of iterations}
\Ensure $\mathbf{C} \in \{1, \ldots, q'\}^{H \times W}$ \Comment{Segmentation map}

\State $\mathbf{P}_{\text{GP}} \gets \text{GlobalModel}(I)$ \Comment{Generate global pseudo-labels}
\State $\theta \gets \text{RandomInit}()$ \Comment{Initialize CNN parameters}

\For{$t = 1$ to $T$}
    \State $\mathbf{x} \gets \text{CNN}(I; \theta)$ \Comment{Local feature extraction}
    \State $\mathbf{R} \gets \text{Conv}_{1 \times 1}(\mathbf{x})$ \Comment{Response map}
    \State $\mathbf{R'} \gets \text{BatchNorm}(\mathbf{R})$ 

    \State $\mathbf{C}_n \gets \arg\max_j \mathbf{R'}_{n,j} \ \forall n \in \{1, \ldots, N\}$ \Comment{Cluster assignment}

    \State \Comment{Compute total loss}
    \Statex \hspace{\algorithmicindent} $L = L_{\text{sim}}(\mathbf{R'}, \mathbf{C}) + \frac{q'}{\alpha} L_{\text{con}}(\mathbf{R'}) + \frac{1}{q'} L_{\text{GP}}(\mathbf{R'}, \mathbf{P}_{\text{GP}})$

    \State $\theta \gets \theta - \eta \nabla_\theta L$ \Comment{Gradient descent}
\EndFor

\State \Return $\mathbf{C}$

\end{algorithmic}
\end{algorithm}

\subsection{DiffSeg and Unsupervised Global Guidance}

DiffSeg serves as the primary external guidance source in DynaGuide, providing high-level semantic information through zero-shot segmentation. It utilizes stable diffusion models to generate segmentation masks from self-attention maps, capturing essential semantic structures across an image without requiring labeled data.

In DynaGuide, DiffSeg produces a set of global pseudo-labels, denoted as $\mathbf{P}_{DS}$, representing the initial segmentation predictions. These pseudo-labels remain static throughout the training process and serve as a reliable external guidance source. By maintaining the global context, they provide a coarse understanding of object boundaries and regions of interest.
The CNN, acting as the local refinement module, leverages these pseudo-labels to iteratively enhance segmentation accuracy. The static nature of DiffSeg’s outputs ensures that DynaGuide maintains robust semantic consistency, preventing overfitting during the refinement process.

While the primary implementation in this study uses DiffSeg for global guidance, DynaGuide’s modular design allows easy integration of other models. Alternatives such as SegFormer or similar supervised segmentation models can also provide pseudo-labels, offering flexibility in adapting the framework to various applications. Crucially, no fine-tuning is performed on the global model, ensuring the segmentation remains fully unsupervised.


\subsection{Local Feature Extraction and CNN Refinement}

The CNN backbone in DynaGuide is designed to capture fine-grained spatial features essential for boundary refinement. Unlike traditional approaches that apply aggressive downsampling, DynaGuide maintains the input resolution throughout the process. This ensures the preservation of spatial information, leading to more accurate boundary segmentation.
The architecture consists of three convolutional blocks, each comprising a convolutional layer, batch normalization and ReLU activation.
To further enhance the learning capacity, residual connections are incorporated, addressing the vanishing gradient problem often encountered in deeper networks. These skip connections provide an additional pathway for gradients to flow, facilitating effective training convergence.

The extracted feature maps, denoted as 
\(\mathbf{x} \in \mathbb{R}^{H \times W \times p}\),
are subsequently processed using a 1x1 convolutional layer with \(q\) filters to produce response maps
\(\mathbf{R} \in \mathbb{R}^{H \times W \times q}\).

The output response maps are normalized using batch normalization to produce the final response map:
\(\mathbf{R'} \in \mathbb{R}^{H \times W \times q}.\)
Following this, each pixel is assigned a cluster label \(\mathbf{C}_n\) using the argmax function:
\begin{equation}
\mathbf{C}_n = \arg\max_j \mathbf{R'}_{n,j}, \quad \forall n \in \{1, \ldots, N\},
\label{eq:argmax}
\end{equation}
where:
\begin{list}{}{}
\item{\( N = H \times W \) denotes the total number of pixels,}
\item{\(\mathbf{C} \in \{1, \ldots, q'\}^{H \times W},\)}
\item{\( q' \) is the dynamically updated number of clusters}
\end{list}
\subsection{Dynamic Dual-Guidance and Loss Function Components}
A key innovation of DynaGuide’s segmentation process is its dynamic loss function, which refines segmentation predictions through adaptive optimization. Unlike static pseudo-labeling approaches that are prone to confirmation bias, DynaGuide’s loss function continuously adapts by aligning local predictions with global guidance. This adaptive nature allows the model to iteratively enhance boundary accuracy and segmentation coherence.
The dynamic loss function comprises three main components:
\begin{itemize}{}{}
\item{\textbf{Feature Similarity Loss} ($L_{\text{sim}}$) promotes the clustering of pixels with similar features.}
\item{\textbf{Spatial Continuity Loss} ($L_{\text{con}}$) enforces smooth and coherent segmentation by applying the Huber Loss to minimize discontinuities in pixel labels.}
\item{\textbf{Global Pseudo-Label Guidance Loss} ($L_{\text{GP}}$) aligns local predictions with the global guidance provided by DiffSeg.}
\end{itemize}
The total loss function is defined as:

\begin{equation}
L = L_{\text{sim}} + \frac{q'}{\alpha} L_{\text{con}} + \frac{1}{q'} L_{\text{GP}},
\label{eq:total_loss}
\end{equation}
where:  \( \alpha \) is a scaling factor used to control the influence of spatial continuity loss.

Crucially, $\alpha$ does not directly balance the loss components. Instead, it modulates the sensitivity of the dynamic weight with respect to the evolving number of predicted clusters \( q' \). This distinction makes the weighing balance more stable and generalizable parameter. To ensure robustness across datasets and eliminate the need for dataset-specific tuning, we fix $\alpha = 15$ in all experiments. This value was chosen based on a limited grid search, balancing boundary sharpness and noise resilience.

\subsubsection{Feature Similarity Loss}
The feature similarity loss \( L_{\text{sim}} \), originally introduced in Differentiable Feature Clustering (DFC)~\cite{kim2020unsupervised}, promotes clustering of pixels with similar features. It is defined as the cross-entropy loss between the normalized response map \( \mathbf{R'}_n \) and the predicted cluster labels \( C_n \).
Mathematically, it is expressed as:
\begin{equation}
L_{\text{sim}}(\mathbf{R'}_n, C_n) = -\sum_{n=1}^{N} \sum_{j=1}^{q} \delta(j - C_n) \ln \mathbf{R'}_{n,j}
\label{eq:feature_similarity_loss}
\end{equation}

where:
\[
\delta(t) = 
\begin{cases} 
1 & \text{if } t = 0, \\
0 & \text{otherwise}.
\end{cases}
\]
This formulation ensures that pixels assigned to the same cluster share similar feature representations, promoting accurate and coherent segmentation results.
\subsubsection{Spatial Continuity Loss}

The spatial continuity loss in DynaGuide builds on the original design introduced in our previous work DynaSeg~\cite{guermazi2024dynaseg}, where it was first formulated as a high-pass filter that encourages spatially continuous pixels to receive the same label by minimizing horizontal and vertical differences in the response map. The initial approach used an \( L1 \) norm-based formulation, which effectively reduced label noise and improved segmentation coherence. However, while effective, this approach had limitations, particularly in preserving spatial consistency for complex structures and textures. To address these limitations, we introduce two key enhancements in DynaGuide:

\begin{itemize}
    \item \textbf{Huber Loss for Enhanced Robustness}: Instead of the \( L1 \) norm, we employ Huber Loss to better manage continuity differences. Huber Loss provides a balance between quadratic and linear penalties, making it more robust to noise and improving segmentation accuracy. This formulation helps in preserving fine-grained details while preventing excessive smoothing of object boundaries.
    \item \textbf{Diagonal Losses for Comprehensive Continuity}: While DynaSeg approach considered only horizontal and vertical pixel relationships, we extend the spatial continuity loss to account for diagonal differences as well. This modification ensures that segmentation remains coherent across all directions, which is particularly useful for handling images with complex textures or diagonal structures. 

    The complete spatial continuity loss in DynaGuide is defined as:
    \begin{equation}
    L_{\text{con}}(\mathbf{R'}_n) = L_{h}(\mathbf{R'}_n) + L_{h}(\mathbf{R'}_n) + L_{d}(\mathbf{R'}_n),    
    \label{eq:con_loss}
    \end{equation}
    \begin{equation}
    L_{h}(\mathbf{R'}_n) = \sum_{i=1}^{H} \sum_{j=1}^{W-1} \text{Huber}(\mathbf{R'}_{i,j+1} - \mathbf{R'}_{i,j}),
    \label{eq:huber_h}
    \end{equation}
    \begin{equation}
    L_{v}(\mathbf{R'}_n) = \sum_{i=1}^{H-1} \sum_{j=1}^{W} \text{Huber}(\mathbf{R'}_{i+1,j} - \mathbf{R'}_{i,j}),
    \label{eq:huber_v}
    \end{equation}
    \begin{equation}
    L_{d}(\mathbf{R'}_n) = \sum_{i=1}^{H-1} \sum_{j=1}^{W-1} \text{Huber}(\mathbf{R'}_{i+1,j+1} - \mathbf{R'}_{i,j}),
    \label{eq:huber_d}
    \end{equation}
    where:
    \begin{itemize}
        \item \( L_{h}(\mathbf{R'}_n) \), \( L_{h}(\mathbf{R'}_n) \) and \( L_{d}(\mathbf{R'}_n) \) captures horizontal, vertical and diagonal continuities using Huber Loss:
        
    \end{itemize}

    By incorporating Huber Loss and diagonal continuity components, DynaGuide improves upon the spatial continuity loss from DynaSeg~\cite{guermazi2024dynaseg}, offering better segmentation robustness, sharper object boundaries, and enhanced adaptability across various datasets.
\end{itemize}

\subsubsection{Global Pseudo-Label Guidance Loss}
The \textit{Global Pseudo-Label Guidance Loss} \(L_{\text{GP}}\) is introduced in DynaGuide to refine segmentation by incorporating high-level global guidance. While DynaSeg and DFC focused solely on feature similarity and spatial continuity, DynaGuide enhances segmentation robustness by integrating pseudo-labels from DiffSeg. These pseudo-labels provide a global segmentation map that complements the local refinements produced by the CNN. 
By leveraging the hierarchical attention-based representations of the global model (DiffSeg or SegFormer), this loss ensures label consistency across large structures while refining details at the pixel level.

The Global Pseudo-Label Guidance Loss is computed as the cross-entropy loss between the CNN-normalized response map \( \mathbf{R'}_n \) and global model's pseudo-labels (DiffSeg) \( \mathcal{P}_{\text{GP}} \):

\begin{equation}
L_{\text{GP}}(\mathbf{R'}_n, \mathcal{P}_{\text{GP}}) = -\sum_{i=1}^{N} \sum_{j=1}^{Q_{\text{GP}}} \delta(j - \mathcal{P}_{\text{GP},i}) \ln \mathbf{R'}_{i,j}
\label{eq:segformer_loss}
\end{equation}

In the following section, we present the experimental setup and evaluate DynaGuide's performance using benchmark datasets.

\section{Experimental Results}
\label{experiment}

We evaluate DynaGuide across three benchmark datasets: BSD500~\cite{martin2001database}, PASCAL VOC2012~\cite{everingham2011pascal}, and COCO~\cite{lin2014microsoft}. The primary evaluation uses DiffSeg as the unsupervised global pseudo-label generator, demonstrating DynaGuide’s effectiveness in a fully unsupervised setting. Additionally, to assess its adaptability, we evaluate DynaGuide using SegFormer-generated pseudo-labels. While SegFormer was originally trained with labeled data, it is used solely as a source of global semantic priors in unseen datasets. DynaGuide applies these pseudo-labels to guide its unsupervised segmentation process without using any ground-truth annotations or fine-tuning on the target datasets. This highlights DynaGuide’s flexibility in leveraging different global guidance models.  The results are presented through quantitative comparisons with state-of-the-art methods, qualitative visualizations, and ablation studies that investigate the impact of key components.

\begin{table*}[t]
\label{tab:results}
\centering

\begin{tabular}{|l|c|c|c|c|c|}
    \hline
    \textbf{Method} & \textbf{BSD500 All} & \textbf{Fine} & \textbf{Coarse} & \textbf{Mean} & \textbf{PASCAL VOC2012} \\
    \hline
    IIC \cite{IIC} & 0.172 & 0.151 & 0.207 & 0.177 & 0.273 \\ 
    k-means clustering & 0.240 & 0.221 & 0.265 & 0.242 & 0.317 \\ 
    Graph-based Segmentation \cite{felzenszwalb2004efficient} & 0.313 & 0.295 & 0.325 & 0.311 & 0.365 \\ 
    CNN-based + superpixels \cite{kanezaki2018unsupervised} & 0.226 & 0.169 & 0.324 & 0.240 & 0.308 \\
    CNN-based + weighted loss, $\mu=5$ \cite{kim2020unsupervised} & 0.305 & 0.259 & 0.374 & 0.313 & 0.352  \\ 
    Self-supervised Multi-view Clustering \cite{fang2021self} & 0.316 & 0.266 & 0.391 & 0.339 & 0.383\\
    Double Clustering with Superpixel Fitting \cite{li2024differentiable} & 0.338 & 0.291 & 0.385 & 0.348 & 0.376 \\
    DynaSeg - SCF~\cite{guermazi2024dynaseg} & 0.330 & 0.290 & 0.407 & 0.342 & 0.396 \\ 
    DynaSeg - FSF~\cite{guermazi2024dynaseg} & 0.349 & 0.307 & 0.420 & 0.359 & 0.391 \\
    \hline
    DiffSeg~\cite{tian2024diffuse} & 0.364 & 0.357 & 0.357 & 0.359 & 0.443 \\
    \hline
    \hline
    \textbf{DynaGuide (DiffSeg)} & \textbf{0.566} & \textbf{0.512} & \textbf{0.523} & \textbf{0.534} & \textbf{0.474} \\
    \textbf{DynaGuide (SegFormer)} & \textbf{0.570} & \textbf{0.553} & \textbf{0.543} & \textbf{0.555} & \textbf{0.481} \\
    \hline
\end{tabular}
\caption{Comparison of mean Intersection over Union (mIoU) for unsupervised segmentation across different granularity levels on the BSD500 dataset and on the PASCAL VOC2012 dataset using DynaGuide with DiffSeg and SegFormer pseudo-labels.}
\end{table*}
\subsection{Experimental Setup}
For our experiments, DynaGuide uses a CNN-based feature extractor consisting of $3$ convolutional blocks, each followed by batch normalization and ReLU activation. This lightweight design was chosen to ensure computational efficiency while maintaining sufficient capacity to capture local spatial details. Preliminary experiments demonstrated that increasing the number of convolutional blocks did not yield significant improvements in mIoU, while reducing blocks led to degraded segmentation accuracy. Therefore, the three-block configuration strikes a balance between accuracy and efficiency. Both the feature dimension $p$ and the cluster dimension $q$ are set to 100 to maintain consistent latent representations. 

The model is optimized using Stochastic Gradient Descent (SGD) with a learning rate of 0.1, momentum of 0.9, and no weight decay. 
In our dynamic loss formulation, the parameter \( \alpha \) plays a distinct role from the static weighting parameter \( \mu \) used in DFC~\cite{kim2020unsupervised}. Unlike \( \mu \), which must be manually tuned for each dataset to balance feature similarity and spatial continuity, \( \alpha \) serves as a dataset-independent scaling factor that regulates the rate of variation in the dynamic balancing weight \( \mu(k) \) throughout training. In this context, the number of active clusters \( q' \) dynamically adapts the loss composition, while \( \alpha \) determines how sharply or smoothly this balance evolves as \( q' \) changes.
Through extensive hyperparameter tuning across multiple datasets, we found that setting \( \alpha\) \(= 15 \) provides the best trade-off between enforcing spatial coherence and allowing flexible cluster formation. \( \alpha \) is fixed across all datasets and experiments. This configuration enhances segmentation accuracy, particularly in challenging scenes with fine object boundaries and diverse textures.
\subsection{Evaluation Metrics}

To evaluate the performance of DynaGuide, which operates without labeled data during training, we establish a correspondence between the model's label space and ground truth categories through a remapping and alignment process. This ensures fair and accurate performance measurement across different datasets.

For the COCO-Stuff dataset, we generate predictions for the validation images and construct a confusion matrix between the predicted labels and the ground truth classes. The Hungarian algorithm is applied to solve the label assignment problem by finding a one-to-one mapping that minimizes assignment costs based on Intersection over Union (IoU) scores. This alignment allows us to calculate the mean Intersection over Union (mIoU) and pixel accuracy (pAcc) across both "stuff" and "thing" categories.

BSD500 provides multiple ground truth annotations for each image, allowing evaluation at different levels of segmentation detail. BSD500 Fine evaluates the model using annotations with the highest segment count, while BSD500 Coarse uses the annotation with the fewest segments. BSD500 All averages results across all available annotations, providing a balanced measure of segmentation quality. The BSD500 Mean is calculated as the average of the All, Fine, and Coarse results, offering a balanced assessment of the model’s overall segmentation capability. This multi-level evaluation provides a comprehensive understanding of DynaGuide’s performance across diverse segmentation scenarios.

For PASCAL VOC2012, we calculate mIoU by matching predicted labels to ground truth segments using the IoU score. The Hungarian algorithm is applied to optimize the label assignment. This metric is then averaged across all annotated segments to provide a robust assessment of segmentation performance.

These evaluation metrics provide a fair and reliable comparison with state-of-the-art methods, capturing both segmentation accuracy and label consistency in various segmentation scenarios.

\subsection{Quantitative Results}
\label{sec:experiments}

We evaluate DynaGuide's segmentation performance using DiffSeg as the primary unsupervised baseline and SegFormer-generated pseudo-labels in a zero-shot manner. Results are reported using mean Intersection over Union (mIoU) and pixel accuracy (pAcc) across BSD500, PASCAL VOC2012, and COCO datasets. Comparative results with state-of-the-art methods are presented.

\subsubsection{BSD500 Evaluation}

Table~\ref{tab:results} summarizes the segmentation performance of DynaGuide on BSD500 across different granularity levels. Using DiffSeg as the primary unsupervised baseline, DynaGuide achieves an mIoU of 0.566 on BSD500 Fine, significantly outperforming the baseline DiffSeg, which attained an mIoU of 0.364. This demonstrates DynaGuide’s ability to accurately capture fine-grained segmentation details.

On BSD500 Coarse, DynaGuide maintains robust performance with an mIoU of 0.512, exhibiting superior adaptability in segmenting broader regions compared to DiffSeg’s 0.357 mIoU. The overall mIoU for BSD500 (All) is 0.523, further highlighting DynaGuide’s consistent effectiveness across varying annotation granularities.

\subsubsection{PASCAL VOC2012 Evaluation}

For PASCAL VOC2012, DynaGuide also demonstrates strong segmentation performance, achieving an mIoU of 0.474, surpassing DiffSeg's baseline result of 0.443. The improved results confirm DynaGuide’s capability to effectively segment complex objects and maintain semantic coherence in challenging real-world scenarios.

\subsubsection{COCO Dataset Evaluation}
The COCO dataset presents a challenging segmentation task due to its large variety of object categories and complex background structures. Tables~\ref{tab:coco_all_results} and \ref{tab:coco_stuff_results} report DynaGuide’s performance using DiffSeg as the global pseudo-label generator. 
\begin{table}[t]
\label{tab:coco_all_results}
\centering
{%
\begin{tabular}{|l|c|c|}
    \hline
    \textbf{Method} & \textbf{Backbone} & \textbf{mIoU All} \\
    \hline
    Modified DC~\cite{caron2018deep} & - & 9.8 \\
    IIC~\cite{IIC} & ResNet18 & 6.7 \\
    PiCIE~\cite{cho2021picie} & ResNet18 & 14.4 \\
    DenseSiam~\cite{zhang2022dense} & ResNet18 & 16.4 \\
    DynaSeg - FSF~\cite{guermazi2024dynaseg} & CNN-Based & 30.51 \\
    DynaSeg - SCF~\cite{guermazi2024dynaseg} & ResNet18 + FPN & 30.52 \\
    \hline
    DiffSeg~\cite{tian2024diffuse} & ViT (UNet) & -- \\
    \hline
    \hline
    \textbf{Ours: DynaGuide (DiffSeg)} & \textbf{CNN-Based} & \textbf{42.18} \\
    \textbf{Ours: DynaGuide (SegFormer)} & \textbf{CNN-Based} & \textbf{52.38} \\
    \hline
\end{tabular}
} 

\caption{Comparison of mIoU for unsupervised segmentation on COCO-All. DynaGuide demonstrates notable improvements with DiffSeg-based global pseudo-label guidance.}
\end{table}
DynaGuide achieves a notable mIoU of 42.18 on COCO-All, demonstrating significant improvements over DynaSeg SCF~\cite{guermazi2024dynaseg} (30.52 mIoU) and other state-of-the-art unsupervised methods. This emphasizes the effectiveness of the dual-guidance strategy in capturing both global context and fine-grained details.

For the more challenging 'Stuff' categories, DynaGuide further improves performance with an mIoU of 45.59 and a pixel accuracy (pAcc) of 73.6, outperforming approaches like DINO + CAUSE-TR~\cite{kim2023causal} (41.9 mIoU, 74.9 pAcc) and DynaSeg FSF (54.10 mIoU, 81.1 pAcc). These results illustrate DynaGuide’s superior ability to ensure spatial consistency and refine boundary details.

\subsubsection{Versatility Evaluation with SegFormer Guidance}

To assess DynaGuide’s versatility, we evaluated its performance using SegFormer-generated pseudo-labels as global guidance. Applied in a zero-shot manner without fine-tuning, although SegFormer is a supervised segmentation network trained with labeled data, it is used solely to generate global semantic priors in this context, maintaining DynaGuide's unsupervised setting. This effectively highlights DynaGuide’s adaptability to different sources of global guidance.

Using SegFormer pseudo-labels, DynaGuide achieved an mIoU of 0.570 on BSD500 All, 0.481 on PASCAL VOC2012, 0.523 on COCO-All, and an impressive 0.552 on COCO-Stuff. These results surpass previous state-of-the-art methods, demonstrating DynaGuide’s superior ability to segment both object-centric and scene-level categories. The model’s consistent performance across diverse datasets highlights its robust adaptability, excelling in capturing fine-grained object boundaries and maintaining spatial coherence without requiring ground-truth labels or fine-tuning.

The quantitative results across BSD500 and PASCAL VOC2012, including comparisons with various state-of-the-art methods, are presented in Table~\ref{tab:results}.
\begin{table}[t]
\label{tab:coco_stuff_results}
\centering
{%
\begin{tabular}{|l|c|c|c|}
    \hline
    \textbf{Method} & \textbf{Backbone} & \textbf{mIoU} & \textbf{pAcc} \\
    \hline
    IIC~\cite{IIC} & ResNet18 & 27.7 & 21.8 \\
    PiCIE~\cite{cho2021picie} & ResNet18 & 31.5 & 50.0 \\
    DenseSiam~\cite{zhang2022dense} & ResNet18 & 24.5 & -- \\
    HSG~\cite{ke2022unsupervised} & ResNet50 & 23.8 & 57.6 \\
    ReCo+~\cite{shin2022reco} & DeiT-B/8 & 32.6 & 54.1 \\
    STEGO~\cite{hamilton2022unsupervised} & ViT-B/16 & 23.7 & 52.5 \\
    DINO + ACSeg~\cite{li2023acseg} & ViT-B/8 & 16.4 & -- \\
    DINO + HP~\cite{seong2023leveraging} & ViT-B/8 & 24.6 & 57.2 \\
    DINO + CAUSE-TR~\cite{kim2023causal} & ViT-B/8 & 41.9 & 74.9 \\
    DynaSeg - SCF~\cite{guermazi2024dynaseg} & CNN-Based & 42.4 & 76.7 \\
    DynaSeg - FSF~\cite{guermazi2024dynaseg} & ResNet18 + FPN & 54.1 & 81.1 \\
    \hline
    DiffSeg~\cite{tian2024diffuse} & ViT (UNet) & 43.6 & 72.5 \\
    \hline
    \hline
    \textbf{Ours: DynaGuide (DiffSeg)} & \textbf{CNN-Based} & \textbf{45.6} & \textbf{73.6} \\
    \textbf{Ours: DynaGuide (SegFormer)} & \textbf{CNN-Based} & \textbf{55.2} & \textbf{82.4} \\
    \hline
\end{tabular}%
}

\caption{Comparison of mIoU and pixel accuracy (pAcc) for 'Stuff' categories on COCO-Stuff. DynaGuide consistently outperforms existing methods using DiffSeg pseudo-labels.}
\end{table}

\begin{table}[t]
\label{tab:computational_efficiency_new}
\centering
{%
\begin{tabular}{|l|c|c|}
    \hline
    \textbf{Method} & \textbf{Total Parameters} & \textbf{GFLOPs} \\
    \hline
    DINO + HP~\cite{seong2023leveraging} & 39.6M & 164.15 \\
    IIC~\cite{IIC} & 4.5M & 17.94 \\
    PiCIE~\cite{cho2021picie} & 23.6M & 4.32 \\
    DenseSiam~\cite{zhang2022dense} & 23.7M & 4.38 \\
    STEGO~\cite{hamilton2022unsupervised} & 86.6M & 67.42 \\
    HSG~\cite{ke2022unsupervised} & 28.0M & 7.13 \\
    DynaSeg (CNN-based)~\cite{guermazi2024dynaseg} & 0.19M & 9.75 \\
    DynaSeg (ResNet-18)~\cite{guermazi2024dynaseg} & 12.0M & 1.84 \\
    \hline
    \textbf{DynaGuide (Ours)} & \textbf{0.11M} & \textbf{6.99} \\
    \hline
\end{tabular}%
}
\caption{Comparison of computational efficiency between DynaGuide and other state-of-the-art segmentation methods. DynaGuide achieves a strong balance between accuracy and efficiency.}
\end{table}

\subsubsection{Computational Efficiency Experiments}

To evaluate the computational efficiency of DynaGuide, we compared its parameter count and floating point operations (FLOPs) with various state-of-the-art segmentation models, as summarized in Table~\ref{tab:computational_efficiency_new}.

DynaGuide demonstrates a clear advantage in computational cost, requiring only 106.4K parameters and approximately 6.99 GFLOPs. Compared to large transformer-based models like DINO + HP~\cite{seong2023leveraging}, which require 164.15 GFLOPs and 39.6M parameters, DynaGuide achieves significantly greater efficiency. This makes it well-suited for real-time and resource-constrained applications.

Furthermore, DynaGuide outperforms CNN-based approaches, offering a more efficient alternative to DynaSeg~\cite{guermazi2024dynaseg}, which uses 193.9K parameters and 9.75 GFLOPs. While models like PiCIE~\cite{cho2021picie} and DenseSiam~\cite{zhang2022dense} have slightly lower FLOPs, they rely heavily on pre-trained backbones, whereas DynaGuide is trained from scratch, demonstrating its robustness and versatility. 
\begin{figure*}[!t]
    \centering
    \includegraphics[width=\linewidth]{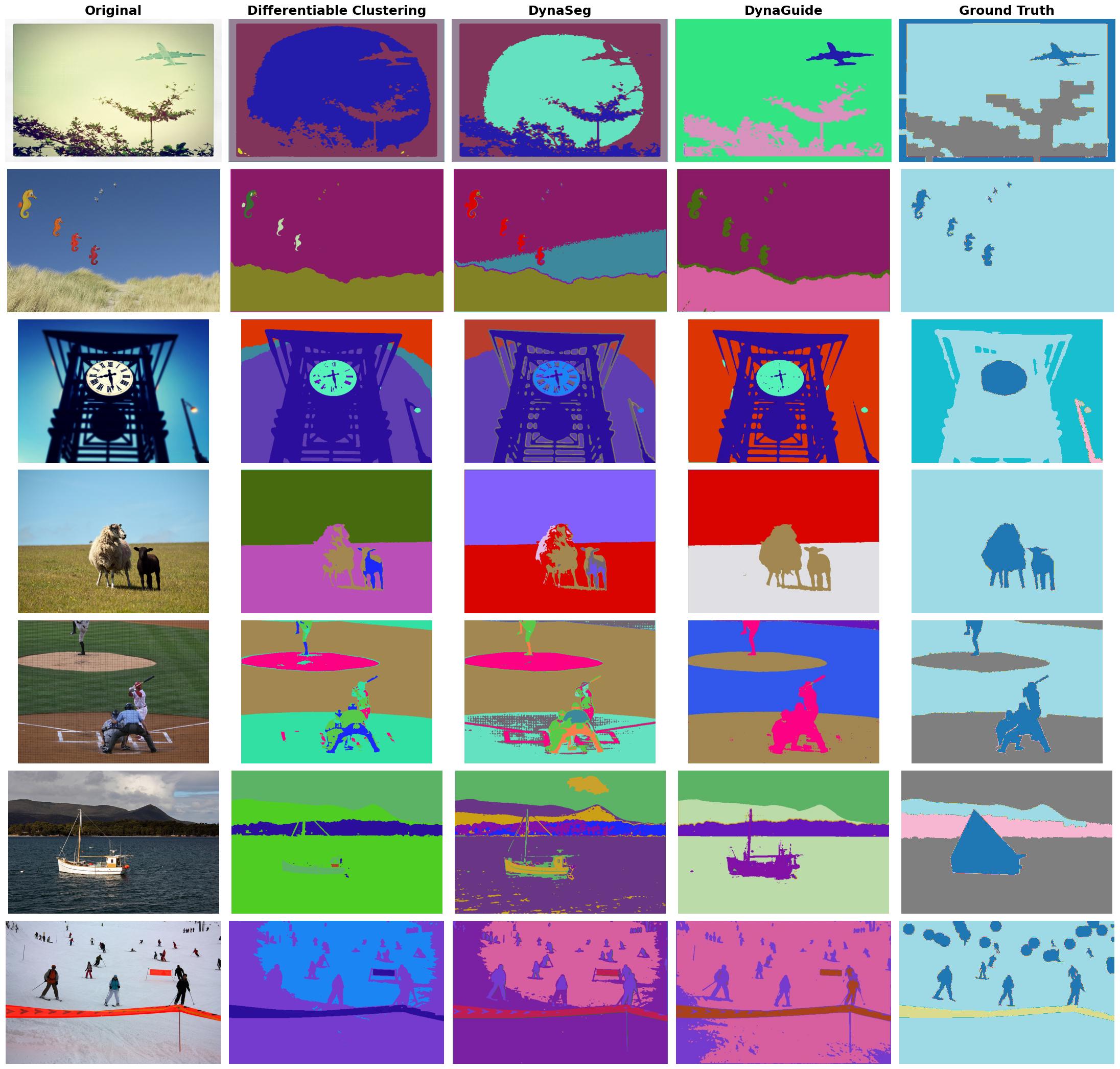}
    \caption{Qualitative comparison of segmentation outputs across different methods: Original Image, Differentiable Clustering~\cite{li2024differentiable}, DynaSeg~\cite{guermazi2024dynaseg}, \textbf{DynaGuide}, and Ground Truth. Detailed examples highlight DynaGuide's ability to address varying brightness, object color differences, and complex shadows.}
    \label{fig:visualizations1}
\end{figure*}

\begin{figure*}[ht]
    \centering
    \includegraphics[width=\linewidth]{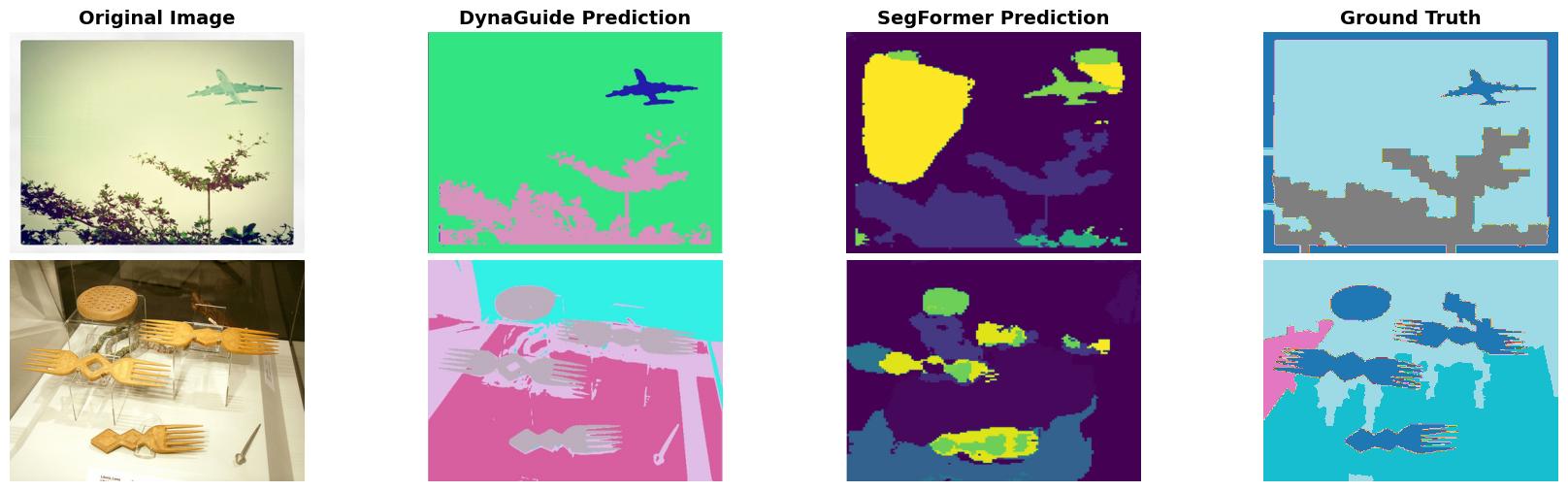}
    \caption{Qualitative comparisons of segmentation outputs for DynaGuide, SegFormer, and ground truth. The first row showcases an airplane scene, and the second row highlights wooden utensils. While SegFormer provides useful attention maps, DynaGuide effectively refines these predictions for enhanced segmentation accuracy.}
    \label{fig:qualitative_comparison}
\end{figure*}
Overall, DynaGuide’s lightweight architecture and minimal computational demands ensure efficient segmentation without compromising accuracy, making it a practical choice for diverse segmentation tasks. DynaGuide’s computational advantages are evident in its low parameter count and FLOPs compared to state-of-the-art methods. With a lightweight CNN architecture and an efficient inference pipeline, DynaGuide achieves significant resource savings. In practical applications, this makes it suitable for deployment on edge devices, mobile platforms, and real-time systems where computational resources are limited. The reduced memory and processing demands translate into faster inference times and lower energy consumption, which is particularly beneficial for large-scale applications in fields like autonomous driving, medical imaging, and video surveillance.

\subsection{Qualitative Evaluation}

\subsubsection{Qualitative Comparison Overview}

Figure~\ref{fig:visualizations1} presents qualitative comparisons of segmentation outputs from DynaGuide, Differentiable Clustering~\cite{kim2020unsupervised}, and DynaSeg~\cite{guermazi2024dynaseg}, alongside the original image and ground truth. These examples underscore the robustness of DynaGuide in tackling diverse segmentation challenges:

\begin{itemize}
    \item \textbf{Airplane Image:} Variations in brightness cause Differentiable Clustering and DynaSeg to undersegment the airplane, blending it with the background. In contrast, DynaGuide produces a distinct segmentation with precise boundaries, achieving refined boundary definitions comparable to the ground truth and offering plausible alternative segmentation results.
    
    \item \textbf{Seahorses Image:} Multiple seahorses with varying colors lead Differentiable Clustering to oversegment the objects. However, DynaGuide accurately segments each seahorse as a cohesive entity while preserving their boundaries.
    
    \item \textbf{Clock Structure Image:} Brightness variations result in oversegmentation by Differentiable Clustering and DynaSeg, creating unnecessary divisions within the clock structure and background. DynaGuide effectively preserves the integrity of the structure with consistent segmentation.
    
    \item \textbf{Sheep Image:} Shadows in the scene cause both Differentiable Clustering and DynaSeg to oversegment the image, misidentifying shadow regions as distinct entities. DynaGuide handles shadows effectively, yielding accurate segmentation of the sheep.
    
    \item \textbf{Skiing Image:} A crowded scene with multiple skiers proves challenging for Differentiable Clustering and DynaSeg, which fail to locate all individuals. DynaGuide, however, segments each skier accurately and achieves precise boundary definitions, even capturing details that the ground-truth mask did not annotate.
\end{itemize}

\subsubsection{Attention Refinement with Global Pseudo-Labels}

One of the key innovations in DynaGuide is the integration of global pseudo-labels (e.g., DiffSeg or SegFormer) to guide CNN-based local segmentation refinement. Pseudo-labels provide strong global priors, which guide the CNN in refining spatial details and achieving boundary precision. This addition ensures that segmentation remains globally coherent, reducing label inconsistencies while maintaining fine details. By combining global pseudo-labels with localized CNN predictions, DynaGuide effectively reduces over-segmentation and enhances fine-grained detail accuracy, particularly in complex scenes.

Figure~\ref{fig:qualitative_comparison} highlights qualitative comparisons between DynaGuide, SegFormer, and ground truth segmentation. Two representative cases are presented:

\begin{itemize}
    \item \textbf{Airplane Image:} While SegFormer effectively captures key regions of the airplane, it struggles with precise boundary delineation. DynaGuide refines these boundaries, producing sharper segmentation results that, in some areas, even outperform the ground truth.
    
    \item \textbf{Wooden Tools:} SegFormer’s attention maps provide strong global cues, but fail to capture the intricate structures of the wooden utensils. DynaGuide’s local refinement mechanism preserves these fine-grained details, ensuring that each object is segmented with greater accuracy.
\end{itemize}

\subsubsection{Limitations and Comparative Analysis with DiffSeg and SegFormer}

\begin{figure*}[ht] \centering \includegraphics[width=\linewidth]{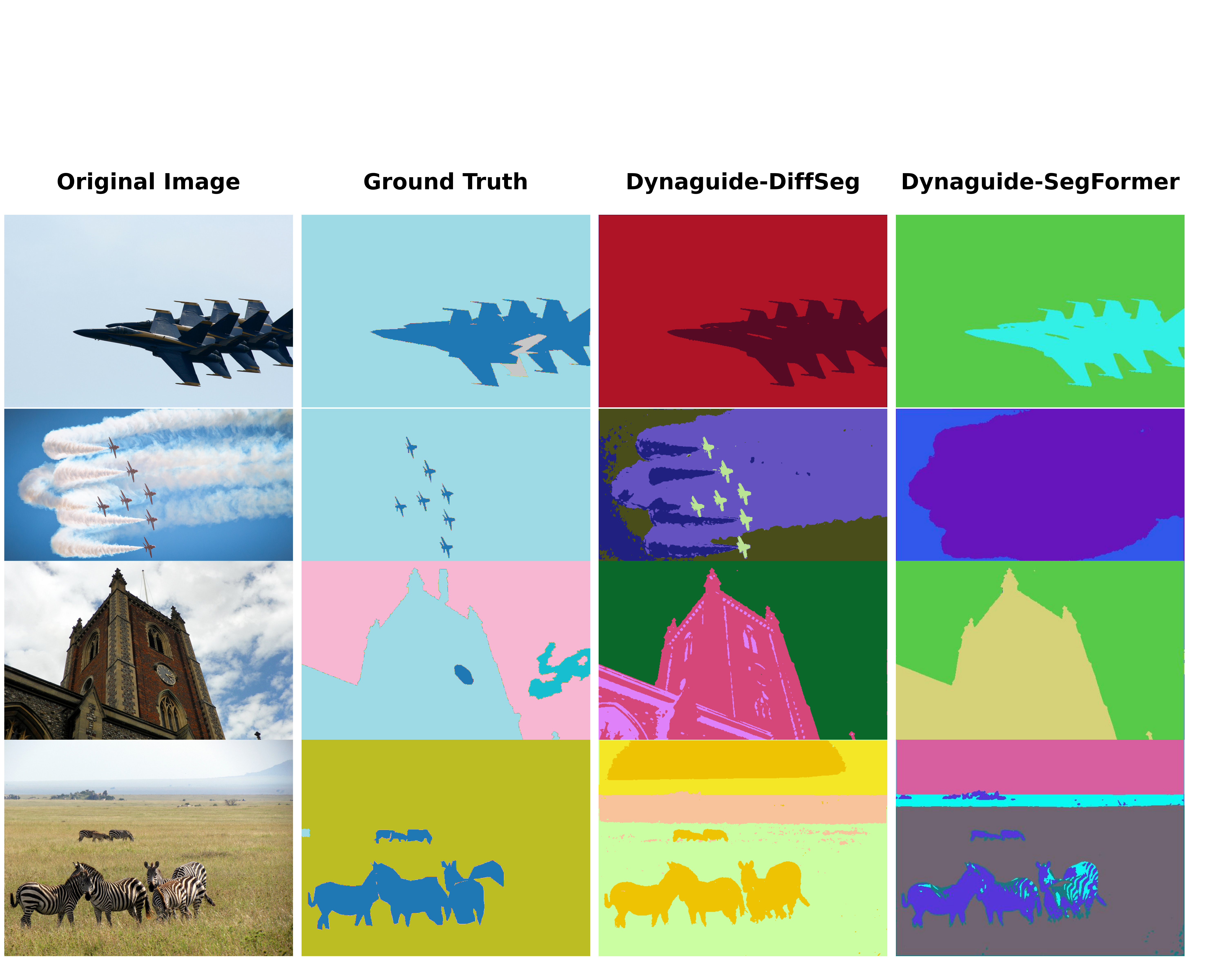} \caption{Qualitative comparisons of segmentation outputs using DynaGuide with DiffSeg and SegFormer pseudo-labels. While both approaches perform well, limitations are evident in certain scenarios. DiffSeg often introduces over-segmentation in complex scenes, while SegFormer may oversimplify details. The figure shows results on airplane formations, airshows, buildings, and landscapes, demonstrating the strengths and limitations of both guidance methods.} \label{fig:visualizations} \end{figure*}

Figure~\ref{fig:visualizations} further compares DynaGuide’s performance using DiffSeg and SegFormer pseudo-labels. While both approaches demonstrate strong results, certain limitations become evident in specific scenarios, especially when DiffSeg is used as the global guidance.

In the airplane formation scene, both DynaGuide versions produce segmentations comparable to the ground truth, accurately identifying the airplanes. However, in the airshow image, DynaGuide with DiffSeg captures not only the airplanes and the sky but also the smoke trails in distinct clusters, demonstrating its sensitivity to texture variations. Conversely, DynaGuide with SegFormer simplifies the segmentation to two clusters, merging the smoke and sky, which results in a loss of finer details.

A more challenging scenario is presented in the building image. The ground truth and SegFormer segment the scene into two regions: the sky and the building, offering a clean and clear segmentation. In contrast, DynaGuide with DiffSeg produces a noisier result, segmenting the building into multiple clusters due to its heightened sensitivity to material differences, shadows, and lighting variations. While this can reveal fine-grained structural details, it also introduces unnecessary fragmentation, demonstrating a limitation in complex scenes with subtle textures.

In the landscape scene with zebras, DynaGuide’s segmentation remains effective, separating both animals and background elements. While both DiffSeg and SegFormer-guided versions perform well, the DiffSeg-guided version introduces minor over-segmentation, particularly in areas with indistinct boundaries, revealing the challenges in maintaining coherence in complex natural scenes.

It is important to note that all results are presented without label matching to the ground truth, as DynaGuide operates in an unsupervised setting. Despite this, its qualitative outputs demonstrate strong segmentation coherence, effective boundary refinement, and adaptability to diverse visual scenarios. By combining global semantic guidance from pseudo-labels with local prediction refinement, DynaGuide shows considerable robustness, though further improvements could address its sensitivity to material variation and texture complexity.

\subsection{Ablation Study}

To validate the contributions of different components in DynaGuide and justify their inclusion in the final architecture, we conduct an ablation study on the BSD500 dataset. By systematically removing or modifying individual elements, we evaluate their impact on segmentation performance using the mean Intersection over Union (mIoU) across all granularity levels.

\subsubsection{Impact of Model Components}

Table~\ref{tab:ablation_loss} presents the results of the ablation study. Replacing the diagonal loss with a standard L1 loss reduces performance, confirming its role in promoting spatial continuity. Removing skip connections results in a notable mIoU drop, demonstrating their effectiveness in enhancing information flow. The absence of global pseudo-label guidance leads to a significant performance reduction, emphasizing the importance of global context. The most severe performance degradation occurs when all three components are removed, underlining their combined effect on accurate segmentation.

These results indicate that diagonal loss aids in maintaining spatial coherence, skip connections mitigate information loss, and global guidance introduces essential semantic context. Their combined effect leads to more accurate and consistent segmentation across different levels of granularity.


\begin{table}[t]
\label{tab:ablation_loss}
\centering
{%
\begin{tabular}{|l|c|c|c|c|}
    \hline
    \textbf{Configuration} & \textbf{All} & \textbf{Fine} & \textbf{Coarse} & \textbf{Mean} \\
    \hline
    DynaGuide (L1 Loss) & 48.77 & 48.36 & 45.53 & 47.55 \\
    No Skip Connection & 56.56 & 53.99 & 53.72 & 54.76 \\
    No Diagonal Loss & 53.74 & 54.57 & 52.83 & 53.71 \\
    No Global Guidance & 50.20 & 50.13 & 45.66 & 48.66 \\
    No Skip, Diagonal, or Guidance & 34.90 & 30.70 & 42.00 & 35.87 \\
    \hline
    \textbf{DynaGuide (Full)} & \textbf{57.96} & \textbf{55.30} & \textbf{54.28} & \textbf{55.84} \\
    \hline
\end{tabular}%
}
\caption{Ablation study results for different configurations of DynaGuide on BSD500, showing mIoU for all granularity levels. The full DynaGuide uses SegFormer as the global pseudo-label generator.}
\end{table}

\subsubsection{Comparison of Feature Extractors}
We further evaluate the effect of using different feature extractors within DynaGuide. Table~\ref{tab:feature_extractors} presents the mIoU results using various architectures, including ResNet18 + FPN, DenseNet, Pointwise DenseNet, and Depthwise Separable Convolutions.

ResNet18 + FPN provides a balanced trade-off between accuracy and computational efficiency, achieving a mean mIoU of 48.73. Depthwise Separable Convolutions demonstrate the best performance among alternative configurations, showing the potential of reducing computational complexity without compromising accuracy. DenseNet and Pointwise DenseNet perform moderately well but are outperformed by DynaGuide’s optimized design.

The results confirm that the combination of diagonal loss, skip connections, and global pseudo-label guidance contributes significantly to DynaGuide’s strong performance. Additionally, the choice of feature extractor further influences the model’s effectiveness, with the complete DynaGuide configuration consistently achieving the highest mIoU scores.

\subsubsection{Visual Impact of Key Components}
\label{sec:visual_ablation}

\begin{figure*}[!htbp]
    \centering
    \includegraphics[width=\linewidth]{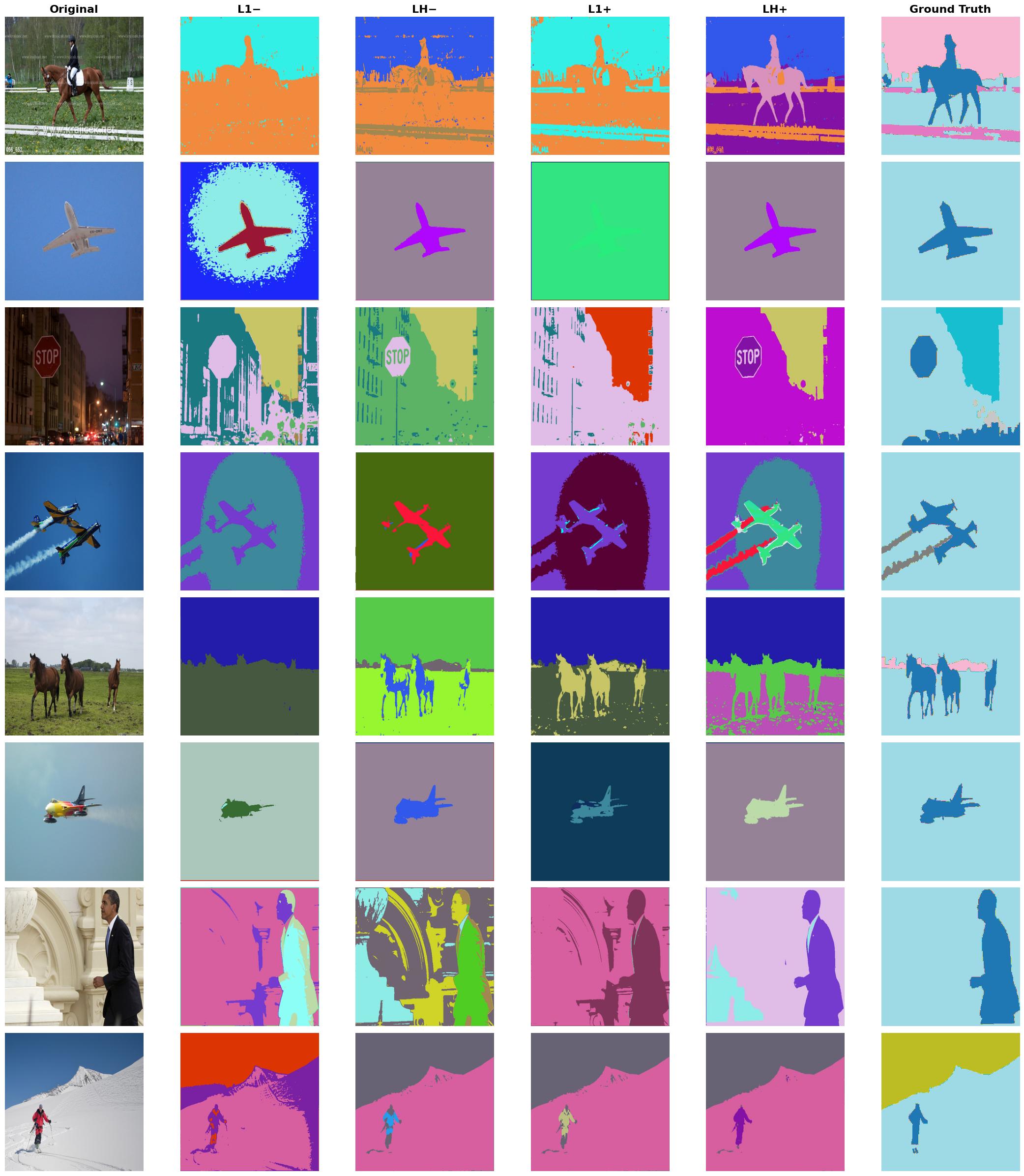} 
    \caption{Qualitative ablation results showing how Huber loss (LH) (Rows 4, 5) and the diagonal component (Row 7) improve segmentation. Configurations are annotated as: \textbf{-} (without diagonal) and \textbf{+} (with diagonal). \textbf{DynaGuide (LH + Diagonal)} provides the most accurate and refined segmentations.}
    \label{fig:ablation_comparison}
\end{figure*}

To further illustrate the role of key design choices in DynaGuide, Figure~\ref{fig:ablation_comparison} presents qualitative comparisons across various configurations. The following observations highlight the impact of the loss function and the diagonal continuity component:

\begin{itemize}
    \item \textbf{Effect of Huber Loss:} Rows 4 and 5 depict scenes with airplanes and horses, respectively. When using L1 loss, these objects are partially merged with the background, resulting in undersegmentation. Specifically, the airplanes in Row 4 are incorrectly labeled as background, while in Row 5, the horses lack clear separation. Replacing L1 with Huber loss resolves these issues, producing more distinct boundaries and preserving object structure.
    
    \item \textbf{Effect of Diagonal Continuity:} Row 7 presents a complex image of Barack Obama. The addition of the diagonal continuity component significantly enhances boundary sharpness and segmentation granularity for both L1 and Huber loss configurations. A similar refinement is visible in Row 5, where the inclusion of diagonal terms results in sharper object boundaries.
    
    \item \textbf{Combined Configuration (Huber + Diagonal):} Across all rows, the full configuration—denoted as \textbf{DynaGuide (LH + Diagonal)}—consistently yields the most accurate results, with clean segment boundaries and minimized under-/over-segmentation.
\end{itemize}

These qualitative results corroborate our quantitative findings and demonstrate that the combined use of Huber loss and diagonal continuity leads to more robust and visually coherent segmentation. The hybrid CNN-Transformer architecture in DynaGuide is thus not only effective in terms of accuracy but also capable of handling complex real-world segmentation scenarios with greater consistency.

\section{Conclusion}
\label{conclusion}
In this paper, we introduced DynaGuide, a novel generalizable dynamic guidance framework for unsupervised semantic segmentation. By leveraging a dual-guidance mechanism that combines global pseudo-labels and local feature refinement, DynaGuide effectively captures both global context and fine-grained details. Our adaptive loss design, incorporating a diagonal continuity loss and multi-component optimization, ensures consistent segmentation results across diverse datasets.

Comprehensive evaluations on benchmark datasets, including BSD500, PASCAL VOC2012, and COCO, demonstrate DynaGuide's robust performance. The framework consistently outperforms prior approaches, achieving higher mIoU and improved segmentation quality. Beyond these results, the model's lightweight architecture and efficient computational footprint make it a practical solution for real-world applications.
\begin{table}[t]
\label{tab:feature_extractors}
\centering
{%
\begin{tabular}{|l|c|c|c|c|}
    \hline
    \textbf{Feature Extractor} & \textbf{All} & \textbf{Fine} & \textbf{Coarse} & \textbf{Mean} \\
    \hline
    ResNet18 + FPN & 48.53 & 48.35 & 47.31 & 48.73 \\
    DenseNet & 47.24 & 44.05 & 43.95 & 45.08 \\
    Pointwise DenseNet & 48.29 & 46.36 & 44.07 & 46.91 \\
    Depthwise Separable & 49.53 & 45.35 & 45.56 & 46.81 \\
    \hline
    \textbf{DynaGuide (Full)} & \textbf{57.96} & \textbf{55.30} & \textbf{54.28} & \textbf{55.84} \\
    \hline
\end{tabular}%
}
\caption{Performance comparison of different feature extractors for DynaGuide on BSD500, showing mIoU for all granularity levels.}
\end{table}

DynaGuide’s versatility extends its potential applications beyond traditional image segmentation. Its capability to adapt to different sources of pseudo-label guidance without relying on labeled data makes it particularly suitable for resource-constrained scenarios. In domains like medical imaging, DynaGuide could enhance diagnosis by segmenting anatomical structures from medical scans. In autonomous driving, the model's efficient inference would support real-time object detection and scene understanding. Furthermore, remote sensing and environmental monitoring could benefit from DynaGuide’s ability to analyze large-scale satellite imagery.

Future research directions include extending DynaGuide to video segmentation tasks, exploring domain-specific adaptations, and further optimizing the loss function for improved robustness. By advancing the field of unsupervised segmentation, DynaGuide offers a scalable and efficient solution for complex visual understanding tasks across a broad range of applications.

 \bibliographystyle{unsrt} 
 \bibliography{refs}

\end{document}